\pdfoutput=1

\documentclass[11pt]{article}

\usepackage[final]{acl}

\usepackage{times, url, booktabs, comment}
\usepackage{latexsym}

\usepackage[T1]{fontenc}

\usepackage[utf8]{inputenc}

\usepackage{microtype}

\usepackage{inconsolata}

\usepackage{graphicx}
\usepackage{multirow}

\title{Exploiting the English Vocabulary Profile for L2 word-level vocabulary assessment with LLMs}

\author{
  \textbf{Stefano Bannò},
  \textbf{Kate M. Knill},
  \textbf{Mark J. F. Gales}
\\
  \textsuperscript{}ALTA Institute, Department of Engineering, University of Cambridge (UK)
\\ 
\texttt{{\{sb2549,kmk1001,mjfg100\}@cam.ac.uk}}
}

\begin{document}
\maketitle
\begin{abstract}


Vocabulary use is a fundamental aspect of second language (L2) proficiency. To date, its assessment by automated systems has typically examined the context-independent, or part-of-speech (PoS) related use of words. This paper introduces a novel approach to enable fine-grained vocabulary evaluation exploiting the precise use of words within a sentence. The scheme combines large language models (LLMs) with the English Vocabulary Profile (EVP). The EVP is a standard lexical resource that enables in-context vocabulary use to be linked with proficiency level. We evaluate the ability of LLMs to assign proficiency levels to individual words as they appear in L2 learner writing, addressing key challenges such as polysemy, contextual variation, and multi-word expressions. We compare LLMs to a PoS-based baseline.  LLMs appear to exploit additional semantic information that yields improved performance.
We also explore correlations between word-level proficiency and essay-level proficiency. Finally, the approach is applied to examine the consistency of the EVP proficiency levels. Results show that LLMs are well-suited for the task of vocabulary assessment.
\end{abstract}

\section{Introduction}

Automated writing evaluation has become an established area of research within natural language processing, playing a key role in language education and computer-assisted language learning~\cite{huawei2023systematic}. Within this context, assessing second language (L2) proficiency remains a critical objective, with increasing attention being paid not only to holistic assessment but also to the evaluation of individual aspects of language competence~\cite{weigle2002assessing}, such as grammar, coherence, and vocabulary. This has proven essential for providing detailed, pedagogically useful feedback to language learners~\cite{hamplyons1995rating}. Among these components, vocabulary is the essential building block of language~\cite{schmitt2001developing} as words are the main vehicle for expressing meaning~\cite{vermeer2001breadth}.

Existing vocabulary assessment methods have typically either assigned an overall vocabulary score to learners’ texts at the essay-~\cite{crossley2023english, banno-etal-2024-gpt} or sentence-level~\cite{arase-etal-2022-cefr}, extracted vocabulary-related features (e.g., lexical diversity or sophistication)~\cite{kyle2015automatically, kyle2018tool}, or attempted to assess the difficulty or appropriateness of individual words~\cite{bax2012text, uchida2018assigning, settles2020machine, aleksandrova-pouliot-2023-cefr}. The latter, however, has been largely under-explored and presents significant challenges -- particularly in dealing with polysemy and contextual variation as well as handling multi-word expressions.

Our work directly addresses this limitation by focusing on word-level, in-context vocabulary assessment in a fully replicable manner. Specifically, we leverage an open-access reference containing information about which lexical items are used at each level of English learning, the English Vocabulary Profile (see Section \ref{sec:evp}). We combine this resource with state-of-the-art large language models (LLMs) to predict the proficiency level of individual words as they are used in L2 learner writing. To the best of our knowledge, this is the first study to apply LLMs for this task, offering a novel and robust approach to vocabulary assessment that explicitly handles semantic ambiguity and contextual nuance.

In Section~\ref{sec:related_work}, we outline the theoretical background of L2 vocabulary assessment and review previous studies that have implemented it in automated systems. Section~\ref{sec:data} presents the English Vocabulary Profile and the L2 datasets used in our experiments. Section~\ref{sec:experiments} constitutes the core of our study. The first experiment focuses on identifying the intended meaning of polysemous words from the English Vocabulary Profile through the use of LLMs, using learner example sentences sourced from the same dataset. In our second experiment, we annotate sentences from a learner corpus by assigning each word a proficiency level based on the English Vocabulary Profile -- annotations we plan to make publicly available. We then automatically predict these levels and compare the performance of various LLMs against a random baseline and a part-of-speech (PoS)-based model. The third experiment extends this approach to additional L2 learner datasets annotated only at the essay level, investigating the correlation between predicted word-level proficiency and essay-level proficiency. Additionally, at the end of the same section, we use our LLM-based approach to examine the consistency of the annotations in the English Vocabulary Profile. Finally, in Section~\ref{sec:conclusions}, we present our conclusions and outline directions for future work.


\section{Related Work}
\label{sec:related_work}
\subsection{Vocabulary assessment}

Despite its fundamental importance, the assessment of vocabulary was only selectively investigated at the beginning of the scientific era of L2 assessment, whereas much more attention was paid to the contrastive analysis of sounds and grammar~\citep{lennon2008contrastive}. When vocabulary knowledge was evaluated, it was primarily tested using the discrete-point approach, an assessment method focused on testing one specific linguistic element -- phonology, morphology, syntax, and vocabulary -- at a time, generally using multiple-choice questions. This approach to vocabulary testing faced various criticisms, as it offered only a limited view of a learner’s vocabulary knowledge, neglected the role of productive language use, disregarded the importance of context in real-world communication, and failed to consider learners' use of strategies to cope with unfamiliar words~\cite{read2000vocab}.

The 1980s represented a watershed in vocabulary assessment since a group of researchers started to publish studies on defined procedures aiming at assessing specific aspects of vocabulary use and knowledge~\citep{anderson1981vocabulary, anderson1982reading, nation1983, meara1987alternative}. These seminal works were something of an exception, given that, on the one hand, the field of L2 acquisition was primarily concerned with the investigation of the acquisition by learners of morpho-syntactic features, whereas, on the other hand, the advent of the communicative approach shifted the attention of language assessment researchers from knowledge of grammatical and lexical elements to the performance of real-world-like tasks~\cite{read2013second}.


For our work, we believe it is important to remember \citeauthor{read2000vocab}'s conceptualisation of vocabulary assessment, who classified it according to 6 dimensions arranged in antonymic pairs: discrete versus embedded, selective versus comprehensive, and context-independent versus context-dependent. The first distinguishes whether vocabulary is assessed as an isolated skill (discrete) or as part of broader language proficiency (embedded). The second refers to the scope of lexical items -- either a specific set (selective) or the learner’s full vocabulary range (comprehensive). The third dimension captures whether vocabulary is assessed in isolation or within authentic contexts. Due to the widespread acceptance of the communicative approach~\cite{harding2014communicative}, it is straightforward to conclude that current trends in language testing and assessment tend to privilege \emph{embedded}, \emph{comprehensive}, and \emph{context-dependent} measures of vocabulary assessment. These three characteristics are central to our approach.

\subsection{Lexical sophistication}

The Common European Framework of Reference (CEFR)~\cite{cefr2001, cefr2020}, a key benchmark aligned with communicative language teaching, testing and assessment, distinguishes between vocabulary range and vocabulary control, which have generally been operationalised along the dimensions of lexical diversity and lexical sophistication, respectively.

Lexical diversity, concerning the breadth of vocabulary used by learners~\citep{yu2010lexical, lu2012relationship}, is typically measured through metrics like type-token ratio or number of unique words. Its relationship with L2 writing proficiency has been widely studied~\citep{crossley2012predicting, gebril2016lexical, treffers2018back, woods2023lexicaldiversity}. While important, lexical diversity is not the primary focus of this paper.



Our work is more closely related to the idea of lexical sophistication. Its focus is the depth of lexical knowledge and is frequently characterised by the presence of relatively rare or uncommon words within a given language sample~\cite{baese2021lexical}. It is generally operationalised using features related to word frequency and familiarity, such as the Lexical Frequency Profile, which reflects the proportion of a learner’s vocabulary falling within various frequency bands derived from a reference corpus~\cite{laufer1995vocabulary}. 
The English Vocabulary Profile (EVP)~\citep{capel2015}, adopted in our work (see Section \ref{sec:data}), is a resource that describes words, phrases, idioms, and collocations used by English learners at different CEFR levels. The study by \citet{lenko2015}, which represents an important precedent for our work, employed it to assign proficiency bands to 90 essays, finding a strong correlation between the clusters obtained using the vocabulary profile and the human-assigned CEFR levels.

Similarly to the EVP, \citet{durlich-francois-2018-efllex} created an online database called EFLLex, which presents the distribution of English words across CEFR levels (A1 to C1), mainly derived from the analysis of textbook corpora. The CEFR-J, a Japan-specific adaptation of the CEFR, also features a word list with proficiency levels~\cite{tono2013cefr}.

\subsection{Automated approaches}
\citet{yoon2012vocabulary} investigated the
use of a vocabulary profile to extract features of lexical sophistication for proficiency assessment of spontaneous speech and found interesting correlations with oral proficiency scores.
\citet{kyle2015automatically} introduced the Tool for Automatic Analysis of Lexical Sophistication (TAALES),
which computes 135 lexical indices. They found that 5 measures of lexical complexity
accounted for more than 50\% of the variance in the human ratings of the spoken and written datasets considered in their study.
Text Inspector~\cite{bax2012text} is an online tool that appears to use the EVP to assess writing proficiency; however, its implementation is not publicly available and is most likely rule-based, as it presents all possible meanings (and corresponding proficiency levels) of a word to the user when ambiguities arise. 
The calculation method used by the CVLA (CEFR-based Vocabulary Level Analyzer) is openly available~\cite{uchida2018assigning}; however, the strictly vocabulary-related part of this tool is also rule-based and, like Text Inspector,
does not appear to address issues related to polysemous or ambiguous words. 

Duolingo developed and released a tool called the CEFR checker, which is now discontinued. This tool allowed users to assess the difficulty level of English and Spanish words and texts. The lexical complexity component of the tool is described in \citet{settles2020machine} and features a vocabulary scale model based on the CEFR framework and a database of 6,823 English words, partly obtained from the EVP. The authors introduce two regression models trained on lexical representations using surface-level features designed to approximate word frequency. However, these models do not appear to account for multi-word expressions or words with multiple meanings.

\citet{gari2021let} demonstrated that BERT could effectively generate contextual embeddings for polysemous words, laying the groundwork for further research in lexical complexity assessment. Building on this, \citet{aleksandrova-pouliot-2023-cefr} explored how the most frequent senses of polysemous words appear in language learner essays. Their study leverages BERT to develop a CEFR-aligned classifier aimed at evaluating the lexical complexity of both single-word and multi-word expressions in English and French. It is important to note that their classifier is trained to predict the CEFR level of an item in context, but not to explicitly identify or disambiguate the meaning or sense of that expression.\footnote{Other works have targeted word-level vocabulary assessment, but not for English~\cite{gala2014modele, alfter-etal-2016-distributions, alfter-volodina-2018-towards}.}

While the aforementioned studies partially or entirely fall short in handling polysemy and ambiguity, our study directly addresses this challenge by targeting word-level, in-context vocabulary assessment in a fully replicable manner. Additionally, to the best of our knowledge, ours is the first study to leverage LLMs for this purpose.

\section{Data}
\label{sec:data}

\subsection{English Vocabulary Profile}
\label{sec:evp}

The English Vocabulary Profile (EVP) \cite{capel2015} is a publicly available reference\footnote{\url{englishprofile.org/}} that contains information about which words, phrases, idioms, and collocations are used at each level of English learning. It is grounded in extensive research using the Cambridge Learner Corpus~\cite{nicholls2003cambridge}, a growing collection of exam scripts written by learners worldwide.

For our experiments, we only considered the British English section of the profile, which includes 15671 entries corresponding to 6747 unique words. The difference arises because some words have multiple meanings or grammatical functions, resulting in several entries comprising multiple options. Additional details about the degree of polysemy are provided in Table \ref{tab:evp_stats} in Appendix \ref{appendix_stats}.

\begin{figure}[!tbhp]
    \centering
    \includegraphics[width=1.0\linewidth]{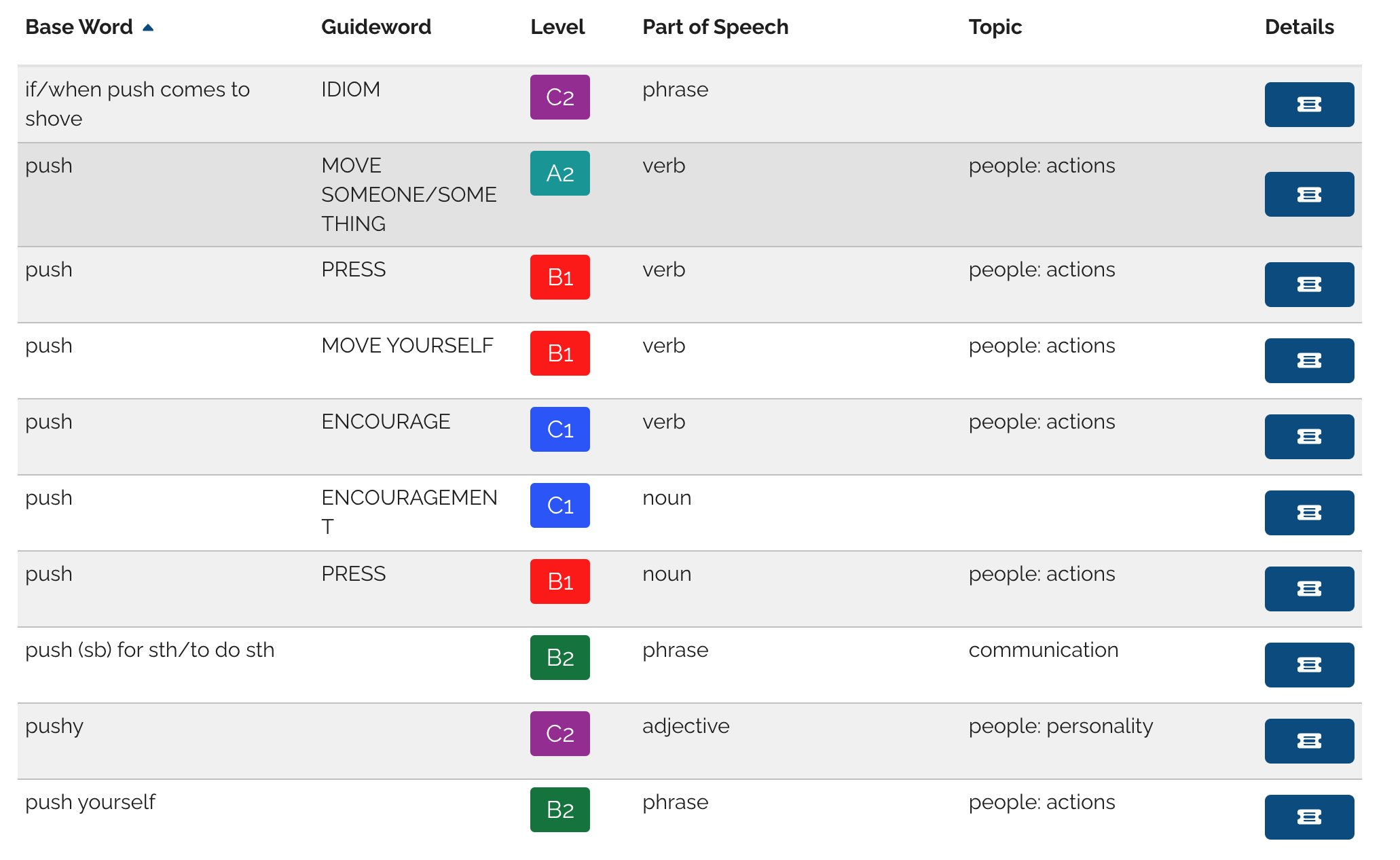}
    \caption{Example for the word \emph{push} from the EVP.}
    \label{fig:evp_example}
\end{figure}

Each EVP entry includes a base word, a guideword, its CEFR level(s), manually assigned PoS, topic, and additional details such as learner and dictionary examples in context (see Figure \ref{fig:evp_example}). Among all entries, when distinguishing by PoS, 62.24\% have more than one CEFR level. However, when considering only unique base words, this proportion drops to 29.21\%. In this work, we refer to words that, within a single PoS, have multiple CEFR levels as \emph{ambiguous} words; the rest are considered \emph{non-ambiguous}. For example, the noun \emph{aim} is classified solely as B1, making it non-ambiguous, whereas the verb \emph{aim} spans levels A2, B2, C1, and C2 depending on context, making it ambiguous.
About 95\% of the section of the EVP we considered in our work has examples taken from L2 learner writing, 
many of which also contain grammatical errors. For the remaining 5\%, we relied on dictionary examples in our experiments.

\subsection{L2 learner datasets}
\label{sec:l2data}

\begin{figure*}[!tbhp]
    \centering
    \includegraphics[width=1.0\linewidth]{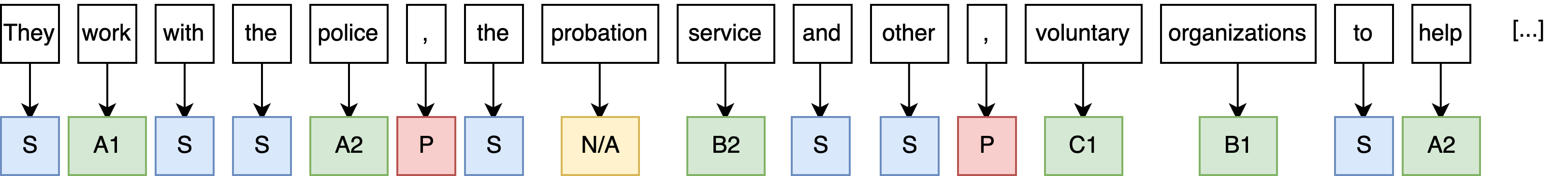}
    \caption{Annotation example on OneStopEnglish. S: stopword; P: punctuation; N/A: word not featured in the EVP.}
    \label{fig:onestop_example}
\end{figure*}


\subsubsection{OneStopEnglish}
OneStopEnglish~\cite{vajjala-lucic-2018-onestopenglish}, is a publicly available corpus\footnote{\url{github.com/nishkalavallabhi/OneStopEnglishCorpus}} for readability assessment and text simplification including 189 parallel compositions across three readability levels: Elementary, Intermediate, and Advanced. We extracted 293 triplets of parallel sentences from the corpus, spanning these three readability levels (see Appendix \ref{sec:appendix_examples} for an example), and annotated them at the word level with CEFR levels from the EVP, excluding stopwords,\footnote{Stopwords include those from the NLTK list (\url{nltk.org}) plus \emph{across}, \emph{among}, and \emph{away}.} punctuation, and words not featured in the EVP (see Figure \ref{fig:onestop_example}). These annotations will be made publicly available.


\subsubsection{EFCAMDAT}

Arguably the largest publicly available L2 learner corpus,\footnote{\url{ef-lab.mmll.cam.ac.uk/EFCAMDAT.html}} the second release of the EF-Cambridge Open Language Database (EFCAMDAT) \citep{geertzen2013automatic, huang2017ef} consists of 1,180,310 scripts written by 174,743 L2 learners as assignments for Englishtown, an online English language school.
The corpus includes 128 distinct writing tasks covering a range of topics, such as describing the rules of a game, reporting a news story, explaining a homemade remedy for fever, and writing to a pen pal.
Each composition is annotated with a proficiency level from 1 to 16, corresponding to CEFR levels A1 to C2.\footnote{The official EF-CEFR mapping can be found at: \url{myenglishlive.ef.com/help-me-article?articleID=55&Vote=Up}}
Learners' first languages are not directly available but can be inferred from their nationalities (approximately 200 in total). For our experiments, we randomly selected 1,000 essays for each CEFR level from the corpus, resulting in a total of 6,000 essays.


\subsubsection{ELLIPSE}

The English Language Learner Insight, Proficiency, and Skills Evaluation (ELLIPSE) Corpus~\cite{crossley2023english} is a publicly available collection\footnote{\url{github.com/scrosseye/ELLIPSE-Corpus}} of approximately 6,500 writing samples from L2 learners of English. Each sample is annotated with both an overall holistic proficiency score and detailed analytic scores covering aspects such as cohesion, syntax, vocabulary, phraseology, grammar, and punctuation conventions. 
Proficiency scores are on scale from 1 to 5. For our experiments, we extracted 359 essays from the training set and 175 from the test set. Further details about the selected essays can be found in Table \ref{tab:ellipse} in Appendix \ref{appendix_stats}.

\section{Experiments}
\label{sec:experiments}

\subsection{Semantic understanding}
\label{sec:semantic_understanding}

Our first experiment serves as a proof of concept and focuses on semantic understanding, with the goal of identifying the intended meaning of a unique word from the EVP when considering the EVP learner examples. Specifically, an LLM is provided with an EVP learner-produced example sentence containing a given word along with all possible EVP entries for that given word. The model's task is to select the most contextually appropriate meaning from the options provided. Note that multi-word expressions are also included by feeding their reference words, e.g., the word \emph{push} for the expression \emph{push (sb) for sth/to do sth} (see Figure \ref{fig:evp_example}). 

The prompt used for this task and a relevant example are reported in Appendix \ref{appendix_prompts} (``Prompt for semantic understanding''). For this experiment, we focused on words with 3, 4, 5, and 6 possible meanings. To avoid positional bias~\cite{liusie-etal-2024-llm}, all possible permutations were considered for words with 3 options.\footnote{In this context, a permutation refers to a unique ordering of the multiple-choice options associated with each question. That is, while the content of the options remains the same, their positions (e.g., labeled A, B, and C) are shuffled across different permutations. This ensures that the model's predictions are not influenced by the fixed position of any particular option, thereby reducing positional bias.} For words with 4, 5, and 6 options, however, only ten permutations were considered due to time and resource constraints. For each permutation, we extract the logits for each option, apply a softmax function, then compute the average probability across permutations and select the option with the highest average probability.

For this experiment, we compared the performance of two proprietary LLMs, i.e., GPT-4o, GPT-4o-mini~\cite{openai2023gpt4}, and three open-source LLMs, i.e., Llama 3.1 8B, Llama 3.1 70B (4-bit quantised) \cite{Llama3}, and Qwen 2.5 32B (4-bit quantised)~\cite{qwen2.5}. These models were selected to ensure a representative range in terms of both model size and the open-source versus proprietary distinction.

Results are evaluated in terms of Accuracy.

\subsection{Word-level proficiency prediction}
\label{sec:wordlevel_exp}

\begin{figure}[!htbp]
    \centering
    \includegraphics[width=1.0\linewidth]{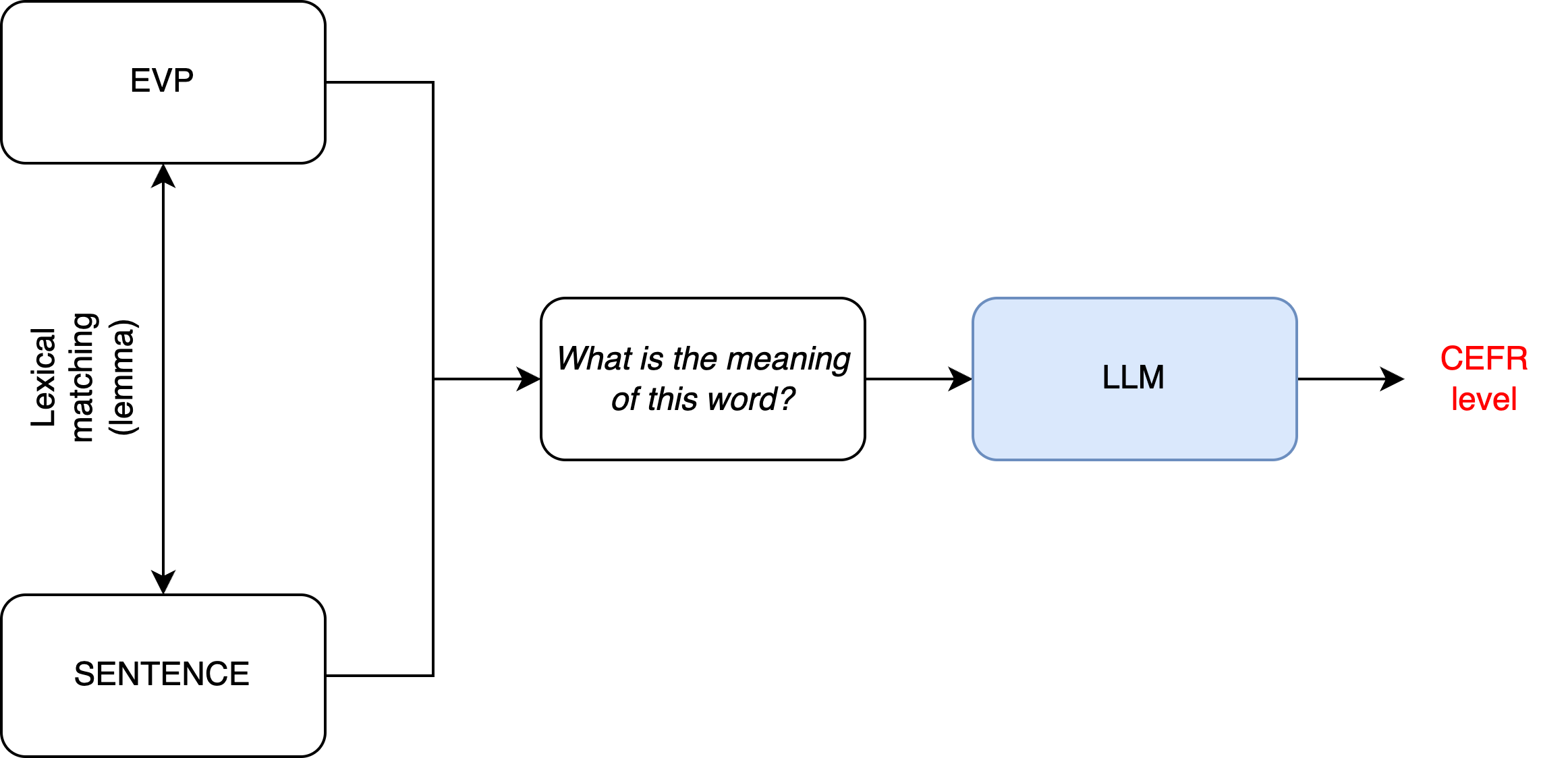}
    \caption{High-level diagram of the approach proposed for word-level CEFR prediction.}
    \label{fig:evp_diagram}
\end{figure}

In the second part of our work, we perform CEFR proficiency level prediction at the word level on the annotated sentences extracted from the OneStopEnglish dataset (see Section \ref{sec:l2data} and Figure \ref{fig:onestop_example}).
We implement the approach proposed in Figure \ref{fig:evp_diagram} in order to extract the proficiency level for each word in a given composition.

Since it is not feasible to feed all EVP entries for each word in a sentence into the LLM at once, we use \texttt{spaCy}\footnote{\url{spacy.io}} to lemmatise each word and map the resulting lemma to its corresponding entries in the EVP. For example, for the word \emph{work} in the sentence shown in Figure~\ref{fig:onestop_example}, the lemma remains \emph{work}, which we then match to all corresponding \emph{work} entries in the EVP. This allows us to filter out irrelevant EVP words in each LLM run. 

\begin{figure}[!htbp]
    \centering
    \includegraphics[width=0.6\linewidth]{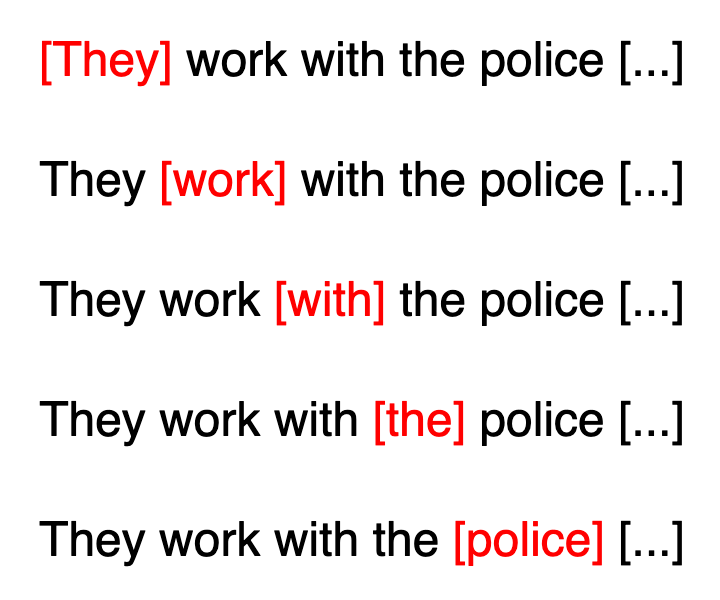}
    \caption{Highlighting method.}
    \label{fig:highlight}
\end{figure}

For each sentence we iterate through each word by highlighting it in square brackets, as shown in Figure \ref{fig:highlight}. If there is no match between a given lemmatised word in the sentence and the EVP, we automatically assign \emph{N/A} to the word (see Figure \ref{fig:onestop_example}). In all the other cases, we add a ``None of the other options'' choice to the other available options and ask the LLM to choose the most suitable option. Then, we extract the logits for each option, apply a softmax function, and select the option with the highest average probability. Finally, we select the CEFR level assigned to this option.
In this set of experiments, we also feed the manually assigned PoS information and a brief definition associated to each option.

The prompt used for these experiments as well as a relevant example can be found in Appendix \ref{appendix_prompts} (``Prompt for word-level CEFR prediction'').\footnote{We also tried this task by prompting the LLMs without EVP information, but results were significantly worse.}

It is important to note that our approach can effectively handle multi-word expressions. For example, when encountering the phrasal verb \emph{take off} in the sense of ``becoming airborne'', the method processes \emph{take} and \emph{off} separately. The word \emph{off} is treated as a stopword and thus excluded from the final assessment. However, \emph{take} is still evaluated in context, allowing the LLM to infer the intended meaning from the surrounding sentence or essay, in addition to the PoS information (i.e., \emph{verb}) and the EVP definition (i.e., \emph{If an aircraft takes off, it leaves the ground and begins to fly.}).
This ensures that the CEFR level assigned reflects the actual usage rather than the isolated form.
Similarly, for the expression \emph{take advantage of sth}, both \emph{take} and \emph{advantage} are mapped to the same corresponding multi-word expression in the EVP, while the stopword \emph{of} is excluded from the evaluation.

For this part of our work, we try the same LLMs as Section~\ref{sec:semantic_understanding} except for GPT-4o-mini and Llama 3.1 8B. We compare the performance of these LLMs with a PoS-based system, which only relies on PoS tags extracted using \texttt{spaCy}. Specifically, if a given lemma in the sentence has only one entry in the EVP 
then its corresponding level is automatically assigned to the word (e.g., \emph{voluntary} in Appendix \ref{sec:appendix_examples} has only one EVP entry, hence one CEFR level). 
Otherwise the relevant PoS tag is matched with the corresponding one in the EVP. If the EVP only specifies one CEFR level for this lemma/PoS pair then this level is  
automatically assigned to the word (e.g., \emph{criminal} in Appendix \ref{sec:appendix_examples} has two entries -- noun and adjective -- at two different CEFR levels). 
Where the pairing is assigned to multiple CEFR levels in the EVP,
the lowest 
of these
CEFR levels is assigned to the word (e.g., the word \emph{service} in the example in Appendix \ref{sec:appendix_examples} has 6 noun entries with 3 different CEFR levels).
Additionally, we consider a random baseline that relies solely on lexical matching. In this case, for words with a single entry, such as \emph{voluntary}, after lemmatisation, we simply assign the associated CEFR level. For words with multiple levels, we randomly pick a level from the available EVP entries.

Results are evaluated in terms of overall Accuracy. Additionally, performance on individual CEFR levels is reported using $F_{1}$ score.

\subsection{Essay-level proficiency prediction}
\label{sec:essay_level}

The third part of our work applies the approach 
presented
in the previous section 
to additional L2 learner datasets, namely EFCAMDAT and ELLIPSE (see Sections~\ref{sec:l2data}). Unlike the OneStopEnglish data, these datasets are annotated only at the essay level and lack word-level CEFR labels. We therefore use our approach to predict the CEFR level of each word and leverage this information as features for essay-level proficiency prediction. For this part of the work, we only used Qwen 2.5 32B\footnote{The prompt is the same reported in Appendix \ref{appendix_prompts} (``Prompt for word-level CEFR prediction'') with the only difference that ``sentence'' is replaced with ``essay''.} and the PoS-based model to extract vocabulary-related information. 
In other words, we aim to explore the predictive power of vocabulary-related features extracted through our models in the context of holistic proficiency assessment. 
Additionally, since the ELLIPSE data include analytic scores targeting specific aspects of proficiency, we further investigate the relationship between these features and scores related to dimensions such as cohesion, syntax, vocabulary, phraseology, grammar, and punctuation conventions.

To do this, for EFCAMDAT, we employ a naive classifier that uses the distribution of predicted CEFR levels within each essay. Specifically, we compute the proportion of words at each predicted CEFR level (i.e., the count of words at each level divided by the total essay length), weight these proportions by their corresponding CEFR levels (i.e., 1 for A1, 2 for A2, 3 for B1, etc.), and sum them to obtain a composite score. We then assess the correlation between this score and the human-assigned holistic score. In addition to the naive classifier, we also employ a simple Support Vector Regression (SVR) model with default parameters (i.e., $\epsilon=0.1$, $C=1$, and a radial basis function kernel). The model is trained using the proportions of words at each predicted CEFR level as input features and evaluated using 5-fold cross-validation.

The SVR is also employed for the ELLIPSE dataset. The model is trained on the training split to predict the holistic proficiency scores. Subsequently, it is evaluated on the test set for both holistic and analytic scores prediction (see Section \ref{sec:l2data}).

Results are evaluated in terms of Pearson's correlation coefficient (PCC) and Spearman's rank coefficient (SRC).

\section*{Experimental results}

\subsection{Semantic understanding results}

Table~\ref{evp_poc_results} reports the results in terms of Accuracy for the semantic understanding task. As can be seen, GPT-4o achieves the best performance, followed by Llama 3.1 70B, with Qwen 2.5 32B performing nearly on par.
While model size appears to play a significant role in the performance gap between GPT-4o and the other models, this pattern does not hold when comparing Qwen 2.5 32B to Llama 3.1 70B, despite the latter being more than twice as large.
All models show a remarkable performance for this task with the exception of Llama 3.1 8B. As expected, Accuracy decreases as the number of options available for a given word increases.

A reasonable question to ask is whether these LLMs have been exposed to the EVP during training, given that it is publicly available. This consideration, among others, motivated us to extend our experiments to additional L2 learner datasets.

\begin{table}[th!]
\small
\centering
\begin{tabular}{l|c|c|c|c|c}
\toprule
\textbf{Model} & \multicolumn{4}{c|}{\textbf{No. of options}} & \textbf{avg.} \\
\cline{2-5}
 & \textbf{3} & \textbf{4} & \textbf{5} & \textbf{6} & \\
\midrule
\textbf{GPT-4o} & {\bf 89.0} & {\bf 86.2} & {\bf 83.1} & {\bf 79.5} & {\bf 84.4} \\
\textbf{GPT-4o$_{mini}$} & 84.4 & 78.9 & 75.4 & 71.2 & 77.5 \\
\textbf{Llama3.1$_{70B}$} & 85.1 & 82.0 & 78.9 & 73.1 & 79.8 \\
\textbf{Llama3.1$_{8B}$} & 77.4 & 70.0 & 64.3 & 64.4 & 69.0 \\
\textbf{Qwen2.5$_{32B}$} & 85.6 & 80.8 & 76.4 & 75.4 & 79.6 \\
\bottomrule
\end{tabular}
\caption{Accuracy (\%) of EVP classification results.}
\label{evp_poc_results}
\end{table}

\subsection{Word-level proficiency prediction results}
\label{sec:wordlevel_results}

\begin{table}[ht!]

\centering
\begin{tabular}{l|c|c|c}
\toprule
\textbf{Model} & \textbf{Ambig.} & \textbf{Non-amb.} & \textbf{All} \\
\midrule
\textbf{Random}   & 29.1  & 88.7  & 61.6\\
\textbf{PoS-based}   & 66.7  & \textbf{93.4} & 80.7 \\
\textbf{GPT-4o}   & 75.1  & 90.5  & 83.3\\
\textbf{Llama 3.1 70B}& 76.9  & 91.5  & 84.6   \\
\textbf{Qwen 2.5 32B}   & \textbf{80.5}  & 92.8  & \textbf{87.0}\\
\bottomrule
\end{tabular}
\caption{Accuracy (\%) results for word-level CEFR prediction (OneStopEnglish).}
\label{tab:accuracy_onestop}
\end{table}

Table \ref{tab:accuracy_onestop} shows the results in terms of Accuracy for the task of word-level CEFR level prediction on the OneStopEnglish data. As mentioned in Section \ref{sec:evp}, we refer to words that have multiple CEFR levels within a single PoS as \emph{ambiguous} words, whereas the rest are considered \emph{non-ambiguous}. We report the results for ambiguous, non-ambiguous, and all words. As expected, the Random classifier performs extremely poorly when predicting the level of ambiguous words. Incorporating PoS information leads to noticeable improvements, as demonstrated by the PoS-based model. However, the best performance on both ambiguous and overall cases is achieved -- in increasing order -- by GPT-4o, Llama 3.1 70B, and Qwen 2.5 32B. Remarkably, Qwen, despite being the smallest of the three LLMs, outperforms all others on this task. The improved performance of LLMs is likely due to their ability to leverage semantic information in addition to grammatical and syntactic knowledge.

For non-ambiguous words, as expected, the Random classifier achieves relatively decent performance. In this setting, the PoS-based model yields the highest accuracy, followed closely by Qwen 2.5 32B. One might expect the PoS-based model to achieve perfect accuracy; however, this is not the case, as the data include instances where words are used with meanings that are not covered in the EVP, leading to misclassifications by the PoS-based model. To address such cases, we included a ``None of the other options'' choice among the LLM's available options (see Section \ref{sec:wordlevel_exp}) -- though this option appears to be over-selected by the model, potentially affecting its accuracy. 

Another reason why the LLMs do not outperform the PoS-based model for non-ambiguous words lies in the way options are selected. When multiple PoS entries exist for a given lemma, the PoS-based model uses only the relevant, non-ambiguous entry matching the POS tag. In contrast, all available options -- including those with irrelevant PoS tags -- are fed into the LLMs. For instance, as illustrated in Section~\ref{sec:evp}, if a sentence contains the word \emph{aim} used as a noun (which appears in the EVP only at the B1 level), the PoS-based model considers only this entry. However, the LLM is presented with both the noun and verb entries (at A2, B2, C1, and C2 in the EVP) for \emph{aim}.\footnote{We deliberately chose not to filter out these entries to avoid assuming perfect accuracy from the \texttt{spaCy} tagger.} These findings are further supported by the performance figures in Table~\ref{tab:accuracy_onestop_breakdown} in Appendix \ref{appendix_other_results}, which reports the breakdown by word-level CEFR level in terms of $F_{1}$ score.

Overall, these results suggest that the most effective solution may lie in a hybrid approach, where an LLM is used to handle ambiguous words, while a PoS-based model deals with non-ambiguous ones.

\subsection{Essay-level proficiency prediction results}
\label{sec:essay_level_res}

\begin{figure}[th!]
    \centering
    \begin{minipage}{0.48\textwidth}
        \centering
        \includegraphics[width=\linewidth]{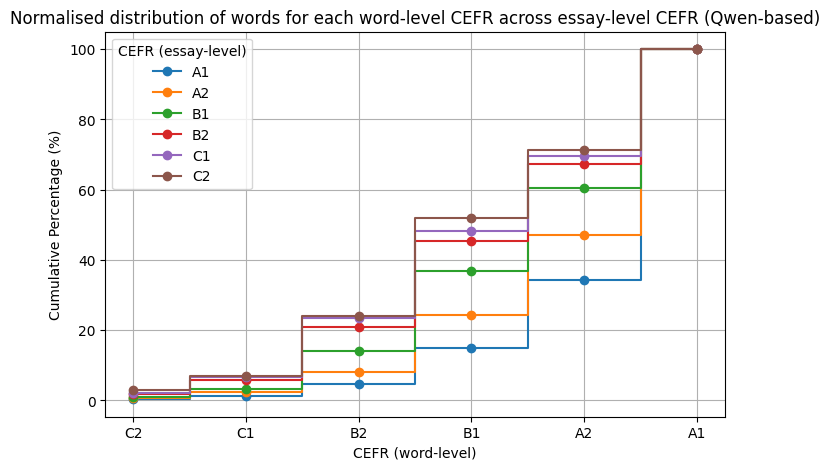}
    \end{minipage}
    \hfill
    \begin{minipage}{0.48\textwidth}
        \centering
        \includegraphics[width=\linewidth]{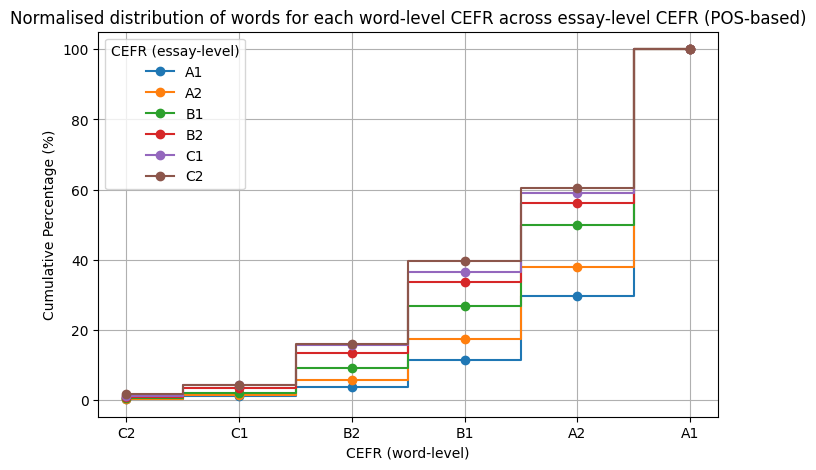}
    \end{minipage}
    \caption{Predicted normalised distribution of words for each word-level CEFR level across essay-level CEFR levels. Qwen 2.5 32B vs PoS-based (EFCAMDAT).}
    \label{norm_distrib_efc_vs_pos}
\end{figure}

We use the same approach to predict word-level CEFR levels on additional L2 learner data, which are only annotated at the essay level. Starting from EFCAMDAT, Figure \ref{norm_distrib_efc_vs_pos} shows a cumulative plot of the predicted normalised distribution of words for each word-level CEFR level across essay-level CEFR levels. We intentionally reverse the order of CEFR levels on the x-axis to emphasise that vocabulary usage is indicative of a certain proficiency level or higher, rather than exclusive to that level. For example, B1-level items are also commonly used by more advanced learners. 

Focusing on a specific element of the figure, if we observe B1 level essays (represented by the green line), we observe that vocabulary from the C2 and C1 levels is used very little, while B2 level vocabulary appears to some extent -- more than in A1 and A2 level essays, but less than in B2, C1, and C2 level essays. Vocabulary from B1 and lower levels is used even more frequently.

\begin{figure}[th!]
    \centering
    \begin{minipage}{0.48\textwidth}
        \centering
        \includegraphics[width=\linewidth]{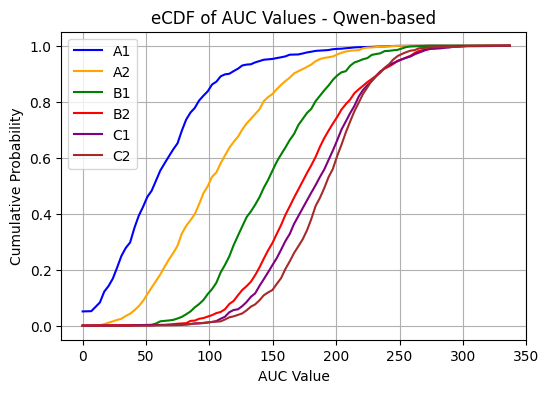}
    \end{minipage}
    \hfill
    \begin{minipage}{0.48\textwidth}
        \centering
        \includegraphics[width=\linewidth]{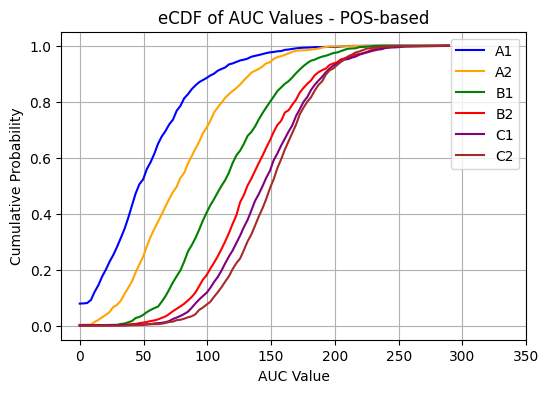}
    \end{minipage}
    \caption{eCDF of AUC values. Qwen 2.5 32B vs PoS-based (EFCAMDAT).}
    \label{ecdf_efc_vs_pos}
\end{figure}

When considering the overall trends, words from A2 to B2 show a steady increase in usage as essay-level proficiency progresses. Interestingly, higher-level words (i.e., C1 and C2) are not as frequently used even in essays written at higher proficiency levels. Nonetheless, a moderate distinction across essay-level proficiency bands can still be observed. To this end, we present Figure~\ref{ecdf_efc_vs_pos}, which displays the empirical cumulative distribution function (eCDF) of the AUC (Area Under the Curve) values computed from Figure~\ref{norm_distrib_efc_vs_pos}. In both figures, we observe larger gaps -- indicating better differentiation across levels -- when using Qwen compared to the PoS-based model.

These findings are further supported by the essay assessment results reported in Table~\ref{pred_power}. When used as features for predicting holistic scores, the vocabulary information extracted with Qwen consistently outperforms that derived from the PoS-based model both in the naive classifier and by the 
SVR.\footnote{These experiments aim to demonstrate the predictive power of the extracted features, not to achieve state-of-the-art essay scoring.}

\begin{table}[ht!]

\centering
\begin{tabular}{cc|c|c}
\hline
Features & Classifier
& \textbf{PCC} & \textbf{SRC} \\
\hline
\textbf{PoS} & \multirow{2}{*}{\textbf{Naive}} & 0.580   & 0.603  \\
\textbf{Qwen} & & \textbf{0.636}   & \textbf{0.656}  \\
\hline
\textbf{PoS} & \multirow{2}{*}{\textbf{SVR}} & 0.734   & 0.713  \\

\textbf{Qwen} &   & \textbf{0.771}  & \textbf{0.749}  \\

\end{tabular}
\caption{Results for essay-level holistic proficiency prediction (EFCAMDAT).}
\label{pred_power}
\end{table}

Finally, we report the results of our experiments on ELLIPSE using Qwen in combination with an SVR trained on holistic scores (see Section \ref{sec:essay_level}).  As shown in Table~\ref{pred_power_ellipse}, despite the high inter-correlation among all analytic scores (see Table \ref{intercorr} in Appendix \ref{appendix_stats}), the SVR (trained on holistic scores) predictions correlate most strongly with the Vocabulary scores, followed by Phraseology, suggesting our features are effectively targeting lexical aspects of language.

\begin{table}[ht!]

\centering
\begin{tabular}{c|c|c}
\hline
& \textbf{PCC} & \textbf{SRC} \\
\hline
\textbf{Overall} & 0.650  & 0.624   \\
\hline
\textbf{Vocabulary} & \textbf{0.637}   & \textbf{0.627}  \\
\textbf{Phraseology} & 0.630 & 0.614 \\
\textbf{Grammar} & 0.577 & 0.556 \\
\textbf{Syntax} & 0.584 & 0.547 \\
\textbf{Cohesion} & 0.595 & 0.569 \\
\textbf{Conventions} & 0.613 & 0.578 \\

\hline

\end{tabular}
\caption{Results for essay-level holistic and analytic proficiency prediction using Qwen-SVR (ELLIPSE).}
\label{pred_power_ellipse}
\end{table}

\subsection{Word-level analysis}

\begin{figure}[ht!]
    \centering
    \begin{minipage}{0.48\textwidth}
        \centering
        \includegraphics[width=\linewidth]{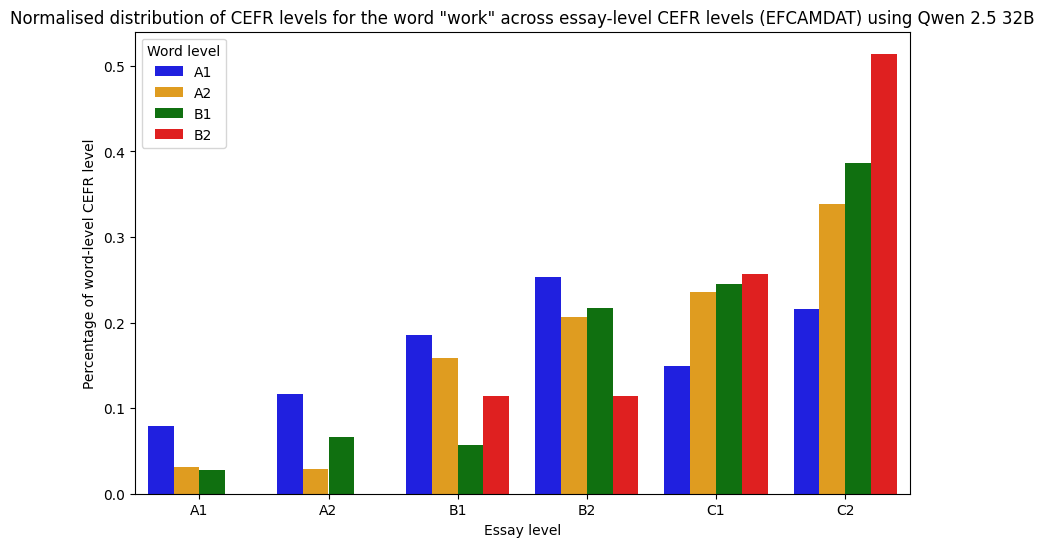}
        \caption{Distribution for word \emph{work} (A1: 191; A2: 278; B1: 482; B2: 667; C1: 457; C2: 667).}
        \label{fig:work}
    \end{minipage}
    \hfill
    \begin{minipage}{0.48\textwidth}
        \centering
        \includegraphics[width=\linewidth]{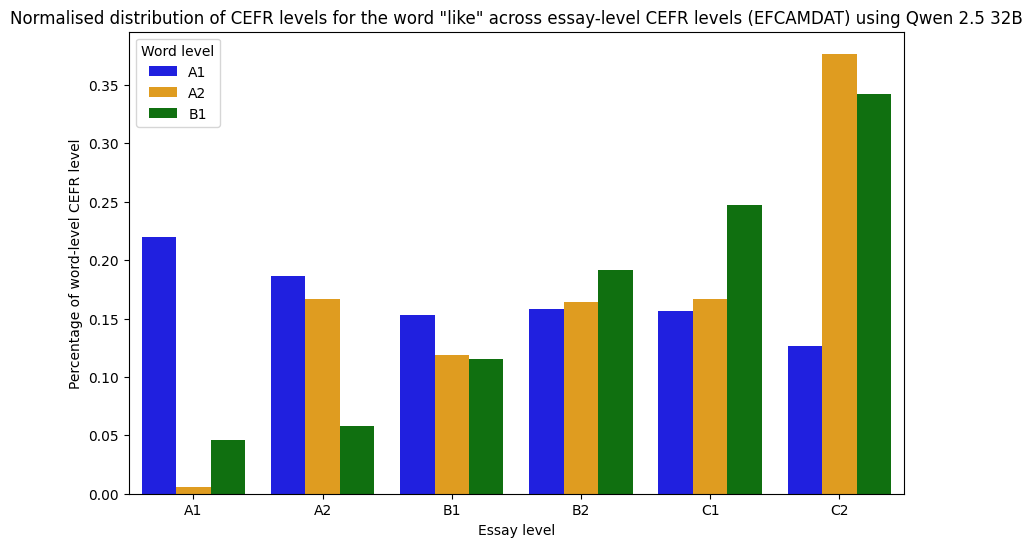}
        \caption{Distribution for word \emph{like} (A1: 401 ; A2: 411; B1: 380; B2: 459; C1: 497; C2: 591).}
        \label{fig:like}
    \end{minipage}
    \label{fig:word_level_analysis}
\end{figure}

Finally, to evaluate the consistency of the EVP, we reverse the approach used thus far. Specifically, we select the two most common words with multiple meanings in the EFCAMDAT data (i.e., \emph{work} and \emph{like}) and examine their word-level CEFR level distribution (predicted with Qwen) across essay-level CEFR levels, as shown in Figures \ref{fig:work} and \ref{fig:like}. The captions report the number of essays containing these words for each essay-level CEFR level.
\begin{table}[ht!]

\centering
\begin{tabular}{c|c|c}
\hline
& \textbf{\emph{work}} & \textbf{\emph{like}} \\
\hline
\textbf{$\geq$B2} & 88.6   & -   \\
\textbf{$\geq$B1} & 90.6 & 89.6 \\
\textbf{$\geq$A2} & 96.8 & 99.4 \\

\hline

\end{tabular}
\caption{EVP consistency in terms of Accuracy (\%) for words \emph{work} and \emph{like} (EFCAMDAT).}
\label{word_level_accuracy}
\end{table}
Figures~\ref{fig:work} and~\ref{fig:like} are summarised in Table~\ref{word_level_accuracy}, where we assess the consistency of the EVP for these two words in terms of Accuracy. As discussed in Section~\ref{sec:essay_level_res}, vocabulary use typically reflects a certain proficiency level or higher, rather than being exclusive to a specific level. Accordingly, we compute Accuracy by checking whether a word appears in essays at its assigned level or above. It is also important to note that we are comparing word-level \emph{vocabulary} levels with essay-level \emph{holistic} scores, where vocabulary represents only one component of the overall assessment. The results show high accuracy and suggest a strong degree of consistency in the CEFR classification provided by the EVP.

\section{Conclusions and future work}
\label{sec:conclusions}

In this work, we introduced a novel approach to in-context, word-level L2 vocabulary assessment by leveraging LLMs in combination with the English Vocabulary Profile. We compared the performance of several open-source and proprietary LLMs to a PoS-based model and showed that the former are particularly effective for this task, especially in handling polysemy and lexical ambiguity.

We plan to integrate this approach into an automatic essay grading system, where it could enrich holistic scoring with fine-grained feedback on vocabulary use, for example, by identifying lexical gaps relative to the learner’s proficiency or highlighting advanced vocabulary usage as a strength. Additionally, we plan to extend our experiments to spoken transcriptions in order to further evaluate the robustness of LLMs and assess their effectiveness across different modalities.

\section*{Limitations}

One limitation of this study is the lack of a systematic investigation into how learner errors affect our approach. In particular, lemmatisation and matching with EVP entries may be hindered by such errors.\footnote{In our first experiment specifically, learner example sentences from the EVP may contain errors, but not involving the target word. Additionally, this experiment does not require lemmatisation.} While spelling mistakes can be addressed with a spellchecker, grammatical or lexical errors pose a greater challenge. In this case, it would be interesting to test our approach on pairs of original and grammatically corrected (manually and/or automatically) sentences or essays and analyse shifts in the LLM's probability distributions in the presence of learner errors.

Another limitation of our approach lies in the way word-level proficiency labels are assigned. In our experiments, each word in the OneStopEnglish data was annotated with its CEFR level, not its specific sense. As a result, polysemous words with multiple meanings but identical CEFR levels could be matched to an incorrect sense, even though the level remains technically accurate. However, it is important to note that the LLMs are prompted to identify the intended meaning of each word based on its context. The CEFR level is then assigned post hoc by mapping that selected meaning to the corresponding entry in the English Vocabulary Profile, hence to a CEFR level.

Finally, due to space constraints, we did not conduct a focused analysis on multi-word expressions such as idioms and phrasal verbs. Nonetheless, given the strong overall results and the fact that multi-word expressions are included in our data, it is reasonable to assume that our approach also performs well on these cases.

\section*{Acknowledgments}

This paper reports on research supported by Cambridge University Press \& Assessment, a department of The Chancellor, Masters, and Scholars of the University of Cambridge. The authors would like to thank the ALTA Spoken Language Processing Technology Project Team for general discussions and contributions to the evaluation infrastructure.


\bibliography{custom}

\appendix

\section{Appendix A: Other stats}
\label{appendix_stats}

Table \ref{tab:evp_stats} shows the degree of polysemy in the EVP.

\begin{table}[ht!]

\centering
\begin{tabular}{lr}
\toprule
\textbf{No. of options} & \textbf{count (\%)} \\

\midrule
1 & 56.96 \\
2 & 21.13 \\
3 & 8.82 \\
4 & 3.99 \\
5 & 2.59 \\
6 & 1.53  \\
$>6$ & 4.98 \\
\bottomrule
\end{tabular}
\caption{Degree of polysemy in the EVP.}
\label{tab:evp_stats}
\end{table}

Table \ref{tab:ellipse} shows the scores distribution in the selected ELLIPSE essays. Note that, in the original corpus, higher (i.e., 4.5 and 5) and lower (i.e., 1 and 1.5) proficiency levels are underrepresented compared to intermediate levels.

\begin{table}[ht!]

\centering
\begin{tabular}{lrrr}
\toprule
\textbf{Overall Score} & \multicolumn{2}{c}{\textbf{\# Essays}} \\
\cmidrule(lr){2-3}
& \textbf{Train} & \textbf{Test} \\
\midrule
5   & 28 & 12 \\
4.5 & 13 & 7  \\
4   & 87 & 43 \\
3.5 & 39 & 23 \\
3   & 61 & 27 \\
2.5 & 62 & 31 \\
2   & 38 & 19 \\
1.5 & 22 & 11 \\
1   & 9 & 2  \\
\bottomrule
\end{tabular}
\caption{Distribution of ELLIPSE essays by overall score in the considered training and test sets.}
\label{tab:ellipse}
\end{table}

Table \ref{intercorr} shows the Correlation between ground truth analytic scores and ground truth overall scores in terms of SRC for the ELLIPSE test split considered in this work.

\begin{table*}[ht!]
\footnotesize
\centering
\begin{tabular}{lrrrrrrr}
\toprule
 & \textbf{Overall} & \textbf{Vocabulary} & \textbf{Phraseology} & \textbf{Cohesion} & \textbf{Grammar} & \textbf{Syntax} & \textbf{Conventions} \\
\midrule
\textbf{Overall} & 1.000 & 0.869 & 0.897 & 0.880 & 0.878 & 0.911 & 0.873 \\
\textbf{Vocabulary} & 0.869 & 1.000 & 0.863 & 0.776 & 0.776 & 0.798 & 0.748 \\
\textbf{Phraseology} & 0.897 & 0.863 & 1.000 & 0.828 & 0.835 & 0.847 & 0.811 \\
\textbf{Cohesion} & 0.880 & 0.776 & 0.828 & 1.000 & 0.781 & 0.825 & 0.816 \\
\textbf{Grammar} & 0.878 & 0.776 & 0.835 & 0.781 & 1.000 & 0.840 & 0.794 \\
\textbf{Syntax} & 0.911 & 0.798 & 0.847 & 0.825 & 0.840 & 1.000 & 0.816 \\
\textbf{Conventions} & 0.873 & 0.748 & 0.811 & 0.816 & 0.794 & 0.816 & 1.000 \\
\bottomrule
\end{tabular}

\caption{Correlation between ground truth analytic scores and ground truth overall scores in terms of SRC.}
\label{intercorr}
\end{table*}

\section{Appendix B: OneStopEnglish examples}
\label{sec:appendix_examples}

The following is an example of three parallel sentences across the three proficiency levels drawn from OneStopEnglish:

\textbf{Elementary:} \emph{They work with the police, the probation service and other, voluntary organizations to help members of the violent criminal gangs of London.}

\textbf{Intermediate:} \emph{They work with the police, the probation service and other, voluntary organizations to help people who feel trapped and frightened in the violent criminal gangs of London.}

\textbf{Advanced:} \emph{They work with the police, the probation service and other, voluntary organizations to help those who feel trapped and frightened in the violent criminal gangs that operate across London.}

\section{Appendix C: Prompts}
\label{appendix_prompts}

\subsection*{Prompt for semantic understanding}

\begin{quote}
\emph{Read this sentence: [\textbf{EVP SENTENCE}] Choose the correct meaning of [\textbf{WORD}] by selecting the most suitable among the following options A, B, C, D, E, or F. No other answer is allowed. Only output the respective option letter without any additional comments, notes, or explanations.}

\emph{A) [DEFINITION A]}

\emph{B) [DEFINITION B]}

\emph{C) [DEFINITION C]} 

\emph{D) [DEFINITION D]}

\emph{E) [DEFINITION E]}

\emph{F) [DEFINITION F]}
\end{quote}

\subsubsection*{Example}

\begin{quote}
\emph{Read this sentence: \textbf{``It was tough on the worn out employees.''} Choose the correct meaning of \textbf{``tough''} by selecting the most suitable among the following options A, B, C, D, E, or F. No other answer is allowed. Only output the respective option letter without any additional comments, notes, or explanations.}

\emph{A) not easy to break or damage}

\emph{B) describes food that is difficult to cut or eat}

\emph{C) Tough people are mentally strong and not afraid of difficult situations.}

\emph{D) difficult}

\emph{E) Tough rules are severe.}

\emph{F) unfair or unlucky}
\end{quote}

\subsection*{Prompt for word-level CEFR prediction}

\begin{quote}

\emph{Read this L2 learner sentence: \textbf{[SENTENCE]}}

    \emph{Choose the correct meaning of \textbf{[WORD]} (in square brackets) by selecting the most suitable among the following options. Also consider the additional information and the PoS of each option. No other answer is allowed. Only output the respective option number without any additional comments, notes, or explanations.}

    \emph{1. [DEFINITION 1] - Additional information: [INFO] (PoS)}

    \emph{2. [DEFINITION 2] - Additional information: [INFO] (PoS)}

    \emph{3. [DEFINITION 3] - Additional information: [INFO] (PoS)}

    \emph{[...]}

    \emph{n. None of the other options.}
    
\end{quote}

where the additional information consists of a brief definition of the word. In round brackets, we also feed the information related to the manually assigned PoS as described in the EVP. See the example below for further information.

The prompt for the experiments conducted on essay-level proficiency prediction is the same. The only difference is that ``sentence'' is replaced with ``essay''.

\subsubsection*{Example}

\begin{quote}

\emph{Read this L2 learner sentence: \textbf{They [work] with the police , the probation service and other , voluntary organizations to help those who feel trapped and frightened in the violent criminal gangs that operate across London .}}

\emph{Choose the correct meaning of \textbf{``work''} (in square brackets) by selecting the most suitable among the following options. Also consider the additional information and the part of speech of each option. No other answer is allowed. Only output the respective option number without any additional comments, notes, or explanations.}
    
\emph{1) the place where you go to do your job - Additional information: work (PLACE) (Part of speech: noun)}

\emph{2) something you do as a job to earn money - Additional information: work (JOB) (Part of speech: noun)}

\emph{3) to do a job, especially the job you do to earn money - Additional information: work (DO JOB) (Part of speech: verb)}

\emph{4) the activities that you have to do at school, for your job, etc. - Additional information: work (ACTIVITY) (Part of speech: noun)}

\emph{5) If a machine or piece of equipment works, it is not broken. - Additional information: work (OPERATE) (Part of speech: verb)}

\emph{6) when you use physical or mental effort to do something - Additional information: work (EFFORT) (Part of speech: noun)}

\emph{7) If something works, it is effective or successful. - Additional information: work (SUCCEED) (Part of speech: verb)}

\emph{8) to exercise in order to improve the strength or appearance of your body - Additional information: work out (EXERCISE) (Part of speech: verb)}

\emph{9) a painting, book, piece of music, etc. - Additional information: work (CREATION) (Part of speech: noun)}

\emph{10) to try hard to achieve something - Additional information: work at sth (Part of speech: verb)}

\emph{11) to spend time repairing or improving something - Additional information: work on sth (Part of speech: verb)}

\emph{12) to do a calculation to get an answer to a mathematical question - Additional information: work sth out or work out sth (Part of speech: verb)}

\emph{13) If a problem or a complicated situation works out, it ends in a successful way. - Additional information: work out (BECOME BETTER) (Part of speech: verb)}

\emph{14) to know how to use a machine or piece of equipment - Additional information: can work sth; know how to work sth (Part of speech: verb)}

\emph{15) to understand something or to find the answer to something by thinking about it - Additional information: work sth out or work out sth (UNDERSTAND) (Part of speech: verb)}

\emph{16) None of the other options}

\end{quote}

\section{Appendix D: Other results}
\label{appendix_other_results}

Table~\ref{tab:accuracy_onestop_breakdown} reports the breakdown by word-level CEFR level in terms of $F_{1}$ score. For ambiguous cases, the PoS-based model performs reasonably well only at the A1 level. However, this result is partly influenced by the rule we set, i.e., assigning the lowest available CEFR level when multiple options are present (see Section \ref{sec:wordlevel_exp}), which inherently favours lower-level classifications. For both the \textit{N/A} and C2 levels, the model yields an $F_{1}$ score of 0, highlighting its complete inability to handle these cases. For non-ambiguous cases, the PoS-based achieves the best performance on all the CEFR levels for the reasons explained in Section \ref{sec:wordlevel_results}, with the exception of the C2 level. For this specific CEFR level, the PoS-based model obtains a high Recall (99.28\%) but a low Precision (73.02\%), while Qwen shows more balanced results, with a Recall of 90.65\% and a Precision of 89.36\%.

\begin{table*}[!ht]
    
    \centering
    \begin{tabular}{l|cc|cc|cc}
        \hline
        & \multicolumn{2}{c|}{\textbf{Ambiguous}} & \multicolumn{2}{c|}{\textbf{Non-ambiguous}} & \multicolumn{2}{c}{\textbf{All}} \\
        & \textbf{Qwen 2.5 32B}  &  \textbf{PoS} & \textbf{Qwen 2.5 32B} & \textbf{PoS} & \textbf{Qwen 2.5 32B} & \textbf{PoS} \\
        \hline
        \textbf{N/A}  & \textbf{50.0}  & 0.0  & 94.3   & \textbf{94.4}  & \textbf{92.0}  & 91.6 \\
        \textbf{A1}   & \textbf{84.4}  & 83.1  & 91.3
        & \textbf{94.0} & \textbf{86.5}  & 85.8\\
        \textbf{A2}   & \textbf{82.1}  & 67.4  & 89.8    & \textbf{91.8} & \textbf{85.1}  & 75.6 \\
        \textbf{B1}   & \textbf{80.6}  & 56.7  & 94.3  & \textbf{95.1}  & \textbf{85.7}  & 72.9\\
        \textbf{B2}   & \textbf{78.5}  & 49.3  & 90.5    & \textbf{90.8} & \textbf{84.2}  & 72.0 \\
        \textbf{C1}  & \textbf{75.5}  & 20.0  & 91.9    & \textbf{92.3} & \textbf{82.9}  & 61.5  \\
        \textbf{C2}  & \textbf{67.2}  & 0.0 & \textbf{90.0}  & 84.2 & \textbf{78.9}  & 59.7 \\
        \hline
    \end{tabular}
    \caption{Breakdown of classification results by word-level CEFR level in terms of $F_{1}$ (OneStopEnglish).}
    \label{tab:accuracy_onestop_breakdown}
\end{table*}

\end{document}